\documentclass[sigconf]{acmart}

\copyrightyear{2022}
\acmYear{2022}
\setcopyright{acmcopyright}\acmConference[MM '22]{Proceedings of the 30th ACM
International Conference on Multimedia}{October 10--14, 2022}{Lisboa, Portugal}
\acmBooktitle{Proceedings of the 30th ACM International Conference on Multimedia
(MM '22), October 10--14, 2022, Lisboa, Portugal}
\acmPrice{15.00}
\acmDOI{10.1145/3503161.3548230}
\acmISBN{978-1-4503-9203-7/22/10}
\usepackage{graphicx}
\usepackage{amsmath}

\usepackage{amssymb}
\usepackage{booktabs}
\usepackage{float}
 \usepackage{pifont}
\usepackage{bbding}
\usepackage{xcolor}
\usepackage{soul}
\hypersetup{
    colorlinks,
    linkcolor={black},
    citecolor={green},
    urlcolor={blue}
}

\usepackage{multirow}
\usepackage{enumitem}

\begin{document}

\title{ARMANI: Part-level Garment-Text Alignment for Unified Cross-Modal Fashion Design}

%%
%% The "author" command and its associated commands are used to define
%% the authors and their affiliations.
%% Of note is the shared affiliation of the first two authors, and the
%% "authornote" and "authornotemark" commands
%% used to denote shared contribution to the research.
\author{Xujie Zhang}
\affiliation{%
  \institution{Shenzhen Campus of Sun Yat-Sen University}
   \city{ShenZhen}
   \state{Guangdong}
   \country{China}
}
\email{zhangxj59@mail2.sysu.edu.cn}

\author{Yu Sha}
\affiliation{%
  \institution{Shenzhen Campus of Sun Yat-Sen University}
    \city{ShenZhen}
   \state{Guangdong}
   \country{China}
}
\email{shay3@mail2.sysu.edu.cn}

\author{Michael C. Kampffmeyer}
\affiliation{%
  \institution{UiT The Arctic University of Norway}
  \state{Tromsø}
  \country{Norway}
}
\email{michael.c.kampffmeyer@uit.no}

\author{Zhenyu Xie}
\affiliation{%
  \institution{Shenzhen Campus of Sun Yat-Sen University}
  \institution{ByteDance}
     \city{ShenZhen}
   \state{Guangdong}
   \country{China}
}
\email{xiezhy6@mail2.sysu.edu.cn}

\author{Zequn Jie}
\affiliation{%
  \institution{Meituan Inc.}
     \city{ShenZhen}
   \state{Guangdong}
   \country{China}
}
\email{zequn.nus@gmail.com}

\author{Chengwen Huang}
\affiliation{
  \institution{Shidi Inc.}
     \city{GuangZhou}
   \state{Guangdong}
   \country{China}
}
\email{huangcw@4dstc.com}
\author{Jianqing Peng}
\authornote{Jianqing Peng is the corresponding author.}
\affiliation{%
  \institution{Shenzhen Campus of Sun Yat-Sen University}
     \city{ShenZhen}
   \state{Guangdong}
   \country{China}
}
\email{pengjq7@mail.sysu.edu.cn}
\author{Xiaodan Liang}
\affiliation{
  \institution{Shenzhen Campus of Sun Yat-Sen University}
     \city{ShenZhen}
   \state{Guangdong}
   \country{China}
}
\email{xdliang328@gmail.com}
\renewcommand{\shortauthors}{Xujie Zhang, et al.}

\begin{abstract}

Cross-modal fashion image synthesis has emerged as one of the most promising directions in the generation domain due to the vast untapped potential of incorporating multiple modalities and the wide range of fashion image applications.
To facilitate accurate generation, cross-modal synthesis methods typically rely on Contrastive Language-Image Pre-training (CLIP) to align textual and garment information. 
In this work, we argue that simply aligning texture and garment information is not sufficient to capture the semantics of the visual information and therefore propose MaskCLIP. MaskCLIP decomposes the garments into semantic parts, ensuring fine-grained and semantically accurate alignment between the visual and text information. Building on MaskCLIP, we propose ARMANI, a unified cross-modal fashion designer with part-level garment-text alignment. ARMANI discretizes an image into uniform tokens based on a learned cross-modal codebook in its first stage and uses a Transformer to model the distribution of image tokens for a real image given the tokens of the control signals in its second stage.
Contrary to prior approaches that also rely on two-stage paradigms, ARMANI introduces textual tokens into the codebook, making it possible for the model to utilize fine-grain semantic information to generate more realistic images.
Further, by introducing a cross-modal Transformer, ARMANI is versatile and can accomplish image synthesis from various control signals, such as pure text, sketch images, and partial images.
Extensive experiments conducted on our newly collected cross-modal fashion dataset demonstrate that ARMANI generates photo-realistic images in diverse synthesis tasks and outperforms existing state-of-the-art cross-modal image synthesis approaches. Our code is available at \url{https://github.com/Harvey594/ARMANI}.

\end{abstract}



\begin{teaserfigure}
  \includegraphics[width=\textwidth]{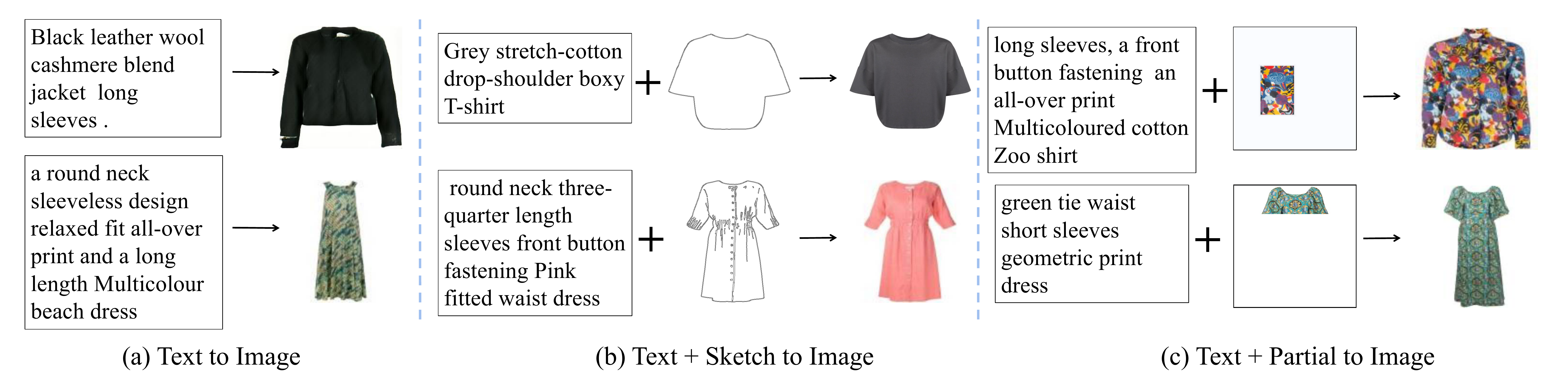}
  \captionof{figure}{
  Our model can generate detailed images from a combination of various cross-modal control signals.} 
  \label{fig:teaser}
\end{teaserfigure}

\maketitle
\section{Introduction}

Cross-modal fashion design, in which a garment image is altered based on control signals from a variety of modalities, such as pure text, sketches, partial images, etc., has the potential to revolutionize the fashion design process. There exist several text-to-image synthesis works~\cite{reed2016generative,dong2017i2t2i,hong2018inferring,li2019object,zhang2018photographic,yuan2018text,huang2018introduction,lao2019dual}, which can be regarded as early attempts of cross-modal fashion design. However, these methods can only achieve text-guided image synthesis, which greatly limits their practicality. Hence, there is increasing demand for a framework that allows the integration of different control signals from diverse modalities for fashion design. However, designing a unified framework for simultaneously handling cross-modal signals is nontrivial, due to their inherent representation differences. 

More specifically, as shown in Fig.~\ref{fig:teaser}, a sketch image is a concise image describing the overall outline of an object, while a partial image is an incomplete RGB image with missing regions. Pure text, on the other hand, differs from the two image counterparts and typically describes a particular object's main characteristics. The representation discrepancy among diverse modalities makes integrating control signals from various modalities difficult for most of the existing cross-modal synthesis works~\cite{xu2018attngan,zhang2018stackgan++,hinz2019semantic,li2020manigan,zhang2021cross,wang2021cycle}, which are usually based on the prevalent generative model, i.e., Generative Adversarial Nets~\cite{reed2016generative}. 
Recently, Vision Transformers~\cite{ding2021,zhang2021ufc,dalle,esser2021taming,zeng2021improving,khan2021transformers,jiang2021transgan,zhu2021tt2inet} have been demonstrated as a feasible paradigm for unified cross-modal image synthesis, where the generated images and the control signals from different modalities are transferred into uniform representations. These methods typically use Vector Quantized Variational AutoEncoders (VQ-VAE)~\cite{oord2017neural} in the first stage to learn a codebook of local features for various visual parts in the real images by compressing the input image into a low-resolution discretized feature map and then reconstructing the input image.
During the second stage, the tokens of the control signals are then fed into a Transformer-based decoder to predict a token sequence for the synthesized image, where predicted tokens are sampled from the learnable codebook in the first stage. 
Benefiting from the global expressivity of the Transformer and the uniform representation for various control signals and the generated image, methods like UFC-BERT~\cite{zhang2021ufc} can address arbitrary types of cross-modal image synthesis within a single model and generate reasonable results that comply with the control signals in most scenarios.
\begin{figure}[tbp]
\centering 
\includegraphics[width=\linewidth]{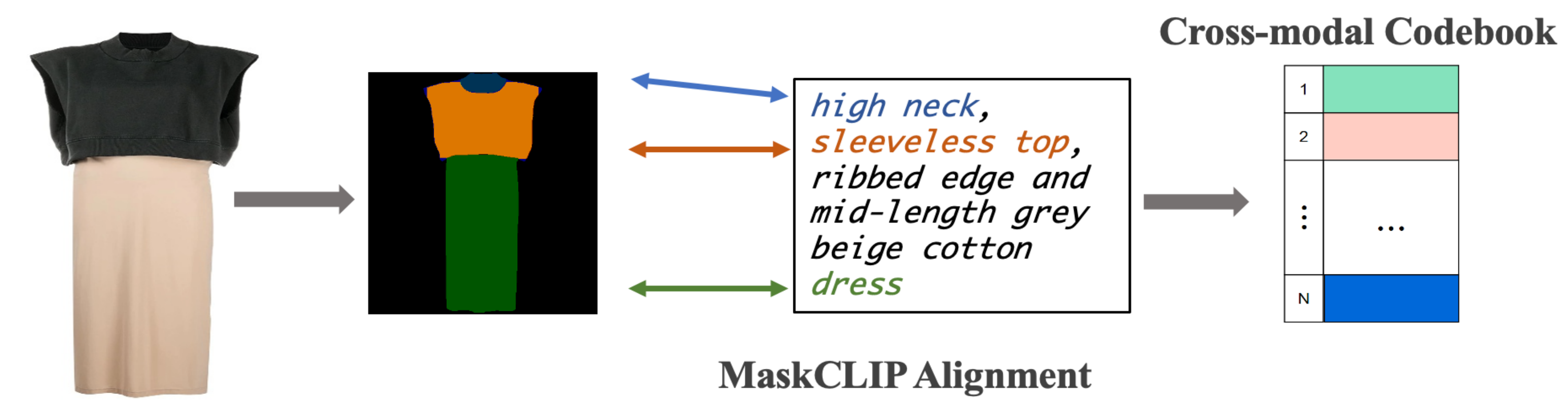} 
\caption{ARMANI leverages a cross-modal codebook, which is facilitated through the proposed MaskCLIP module and its fine-grained and semantically accurate alignment, increasing its expressivity over prior approaches.}
\vspace{-5mm}
\label{1} 
\end{figure}

However, the codebook generation mechanism used by the aforementioned two-stage paradigm inevitably leads to an issue that compromises the generalization of the model.
Since the codebook is learned by reconstructing images from the training set, the codebook can only contain visual information and represent local features of the various visual parts of the training images. Moreover, when testing, generation according to words that appear less frequently in the training set may not be accurate. For instance, most garment descriptions only provide an overall description and do not contain detailed descriptions of the garment components such as buttons. This again will lead to poor generation results.

The prime reason for this issue is that codes in the codebook only encode the local visual features of the images and neglect the corresponding text information that provides more distinguishing features. This leads to an inferior representation with coarse semantics. To address this, text information with more fine-grain semantics should be considered to enhance codebook expressivity. However, naively aligning visual and text information using Contrastive Language-Image Pre-training (CLIP~\cite{radford2021learning}) would be insufficient as it neglects the semantic information of the garment.

In this paper, we therefore propose a pARt-level garMent-text Alignment for uNIfied cross-modal fashion design (ARMANI) to accomplish diverse manipulations according to control signals from different modalities. In contrast to existing two-stage methods, ARMANI divides the first stage (i.e., the codebook learning stage) into two steps. The first step consists of a Mask-level Contrastive Language-Image Pretraining (MaskCLIP) module that decomposes the garments into semantic parts to facilitate fine-grained alignment between the visual information and the text description (illustrated in Fig.~\ref{1}). In the second step, each code in the learnable codebook is updated by collaboratively considering the visual feature and its most related word embedding, resulting in a text-aware and semantically accurate, cross-modal codebook.

The textual semantics in the cross-modal codebook enable the model to predict more reasonable codes for control signals not seen in the training data. Let us consider the example of "a T-shirt with a zip". 
Despite the lack of a specific code for "a T-shirt with a zip", the model can now pay more attention to the textual semantics of the word "zip" and predict the code by leveraging the word embedding of "zip" from other zip items such as "a jacket with a zip".

Furthermore, to increase the versatility of ARMANI, we design a cross-modal Transformer in the second stage, which simultaneously takes as inputs different types of control signals (i.e., pure text, sketches, partial images) to predict the final token sequence. Further, we propose a semantic loss that leverages our proposed MaskCLIP in both stages of the framework to ensure that the cross-model codebook and cross-modal Transformer are semantically accurate.
Finally, we collect a new cross-modal fashion dataset (CM-Fashion) that contains over 500,000 garment images with corresponding textual descriptions. Extensive experiments on our newly collected dataset demonstrate that ARMANI can conduct various fashion manipulation tasks in compliance with diverse control signals (see Fig.~\ref{fig:teaser} and is superior to existing state-of-the-art cross-modal image synthesis works. In summary, our contributions are:

\begin{itemize}[itemsep=0.5pt,topsep=1pt]
\item ARMANI, a novel and versatile framework for cross-modal fashion design that can perform fashion manipulation based on cross-modal control signals and leverages an expressive cross-modal codebook to incorporate the fine-grain text information. 
\item MaskCLIP, a Mask-level Contrastive Language-Image Pretraining module that learns a semantically accurate alignment between the visual and textual features. MaskCLIP further ensures semantic consistency when training the cross-modal codebook and the cross-modal Transformer.
\item CM-Fashion, the largest publicly released cross-modal fashion dataset that both contains high-resolution images from a large variety of garment categories and detailed noise-free text descriptions. 
\item An extensive evaluation of our proposed ARMANI, which demonstrates its benefits over prior cross-modal approaches.

\end{itemize}

\begin{figure*}
\centering 
\includegraphics[width=0.95\textwidth]{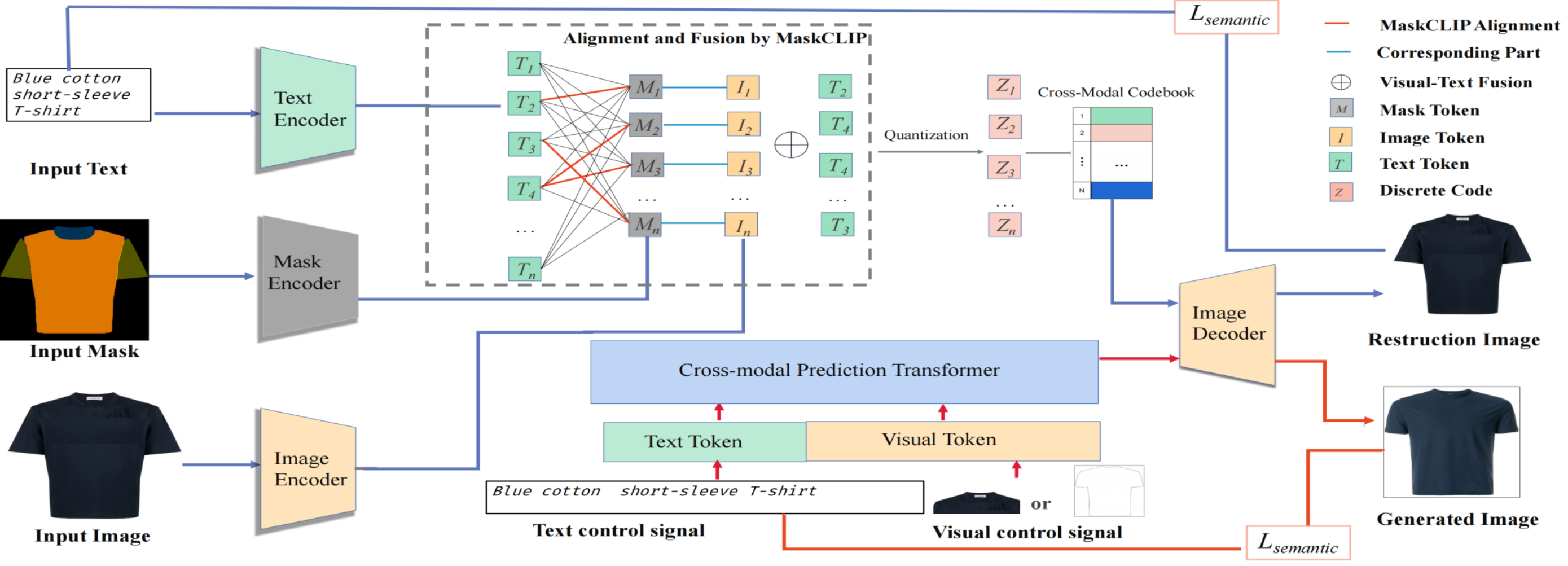}
\caption{The overall architecture of ARMANI, our two-stage framework for cross-modal fashion design. In the first stage, an image, mask and text encoder are used to embed the image, the mask, and the text, respectively. An image decoder is used to generate the garment and to facilitate training via image reconstruction. Mask and text tokens are aligned using MaskCLIP and the most relevant text for a mask is fused with the image token that corresponds to the mask token. After fusion, the token is quantized to obtain z. Finally, a cross-modal codebook is obtained after training. In the second stage, the text control signal and visual control signals are used as inputs to generate a sequence of tokens that is used to generate the image. In addition, we propose a semantic loss based on MaskCLIP, which is used in both stages to improve the semantic awareness of our model.
% In the first stage, an image encoder, a mask encoder, and a text encoder are used to embed the image, the mask, and the text, respectively. An image decoder is used to generate the garment and to facilitate training via image reconstruction. The mask and text tokens are aligned using MaskCLIP and the most relevant text for a mask is fused with the image token that corresponds to the mask token. After fusion, the token is quantized to obtain z. Finally, a cross-modal codebook is obtained after training. In the second stage, the text control signal and visual control signals are used as inputs to generate a sequence of tokens that is used to generate the image. In addition, we propose a semantic loss based on MaskCLIP, which is used in both stages to improve the semantic awareness of our model.
}
\label{2}
\vspace{-4mm}
\end{figure*}

\section{RELATED WORK}
\subsection{Cross-modal Image Synthesis}
Recently, a great deal of progress~\cite{xia2020,StyleGAN,dalle,esser2021taming,zhang2021ufc,li2019connecting,cai2019towards,zhu2020cookgan} has been made on the cross-modal image synthesis task, which aims to generate realistic pictures from various control signals, such as pure text, sketch images, or partial images. Existing methods can be divided into two categories, in which the first category generates images via GAN-based approaches. For instance, TediGAN~\cite{xia2020} learns a StyleGAN~\cite{StyleGAN} inversion module and then computes the visual-linguistic similarity to generate an image from cross-modal inputs. The second category of methods produces images via a two-stage image synthesis framework, where the first stage trains a VQVAE~\cite{oord2017neural} to convert an image into a sequence of discrete tokens and convert such a sequence of tokens back into an image. During the second stage, various control signals are used as inputs and a Transformer is trained to capture the distribution of the token sequences. Leveraging this two-stage framework, DALL·E~\cite{dalle} and Cogview~\cite{ding2021} generate high-resolution images from text descriptions, while M6-UFC~\cite{zhang2021ufc} adopts a BERT-based two-stage framework to unify any number of cross-modal control signals for conditional image synthesis. These methods, however, use discrete tokens without textual information, resulting in sub-optimal generation capabilities. ARMANI addresses this problem by integrating textual information in the first stage, leading to more expressive tokens.
\subsection{Cross-Modal Interaction Mechanism}

One of the key challenges in cross-modal research is the modeling of the interaction between two modalities, which is mainly addressed by two types of approaches: single-stream and dual-stream models. Single-stream models like VisualBERT~\cite{li2019visualbert}, UNITER~\cite{chen2020uniter}, and ViLT~\cite{kim2021vilt} directly concatenate the patch-wise and textual embeddings and leverage a Transformer to model interactions, while dual-stream models like ViLBERT~\cite{li2019visualbert}, ALIGN~\cite{jia2021scaling}, UNIMO~\cite{li2020unimo}, CLIP~\cite{radford2021learning}, FILIP~\cite{yao2021filip}, and GLIP~\cite{li2021grounded} learn separate encoders for the different modalities prior to aligning the modalities, which allows flexible use of different models for different modalities and to address various downstream tasks. Inspired by these dual-stream models, ARMARNI introduces the novel MaskCLIP module to learn the correspondence between the fine-grained features of different modalities.

%CLIP~\cite{radford2021learning} and ALIGN~\cite{jia2021scaling} encode each image or text into a global feature representation and perform cross-modal contrastive learning which aligns the textual and visual information into a unified semantic space. To take into account the fine-grained interaction between image patches and textual tokens, FILIP~\cite{yao2021filip} adopts a contrastive learning objective equipped with cross-modal late interaction to achieve fine-grained alignment and GLIP~\cite{li2021grounded} presents a grounded language-image pretraining model for learning object-level, language-aware, and semantic-rich visual representations. Inspired by these dual-stream models, ARMARNI introduces the novel MaskCLIP module to learn the correspondence between the fine-grained features of different modalities.

\section{ARMANI}
We first briefly discuss the technical details of the two-stage synthesis paradigm, which our model ARMANI (see Fig.~\ref{2}) inherits. We then introduce the components of our framework in Sec.~\ref{sec:maskClip}-\ref{sec:transformer}.

\begin{figure*}
\centering 
\includegraphics[width=0.95\textwidth]{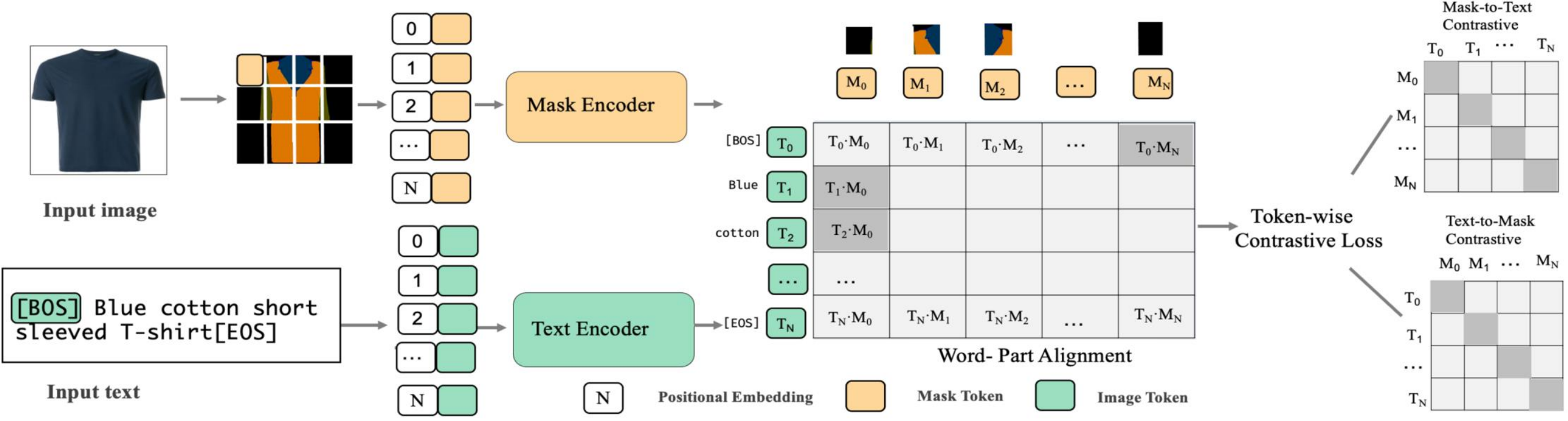}

\caption{Overall architecture of MaskCLIP.The representations of the text and mask tokens are obtained using the mask encoder and the text encoder, respectively. The token-wise contrastive loss is then optimized to learn the word-part alignment.
}
\label{3}
\vspace{-4mm}
\end{figure*}

\subsection{Preparatory: Two-stage Synthesis Framework}
\label{sec:prep}

The first stage aims to learn a semantically rich codebook and an expressive generator, where the codebook encompasses a variety of visual features from the dataset, while the generator can synthesize realistic images given a collection of codes sampled from the codebook. 
This is achieved by introducing a Variational Autoencoder (VAE) with a convolutional encoder (E) and a decoder (G) that is trained to conduct image reconstruction.
Concretely, the encoder $E$ first compresses the input image $\mathbf{I} \in \mathbb{R}^{H \times W \times 3}$ into a low-resolution feature map $\mathbf{\hat{Z}} \in \mathbb{R}^{h \times w \times n_z}$. Then, for each spatial position (i,j) of $\mathbf{\hat{Z}}$, the vector $\hat{z}_{i,j}  \in \mathbb{R}^{n_z}$ is quantized into a discrete code $z_{i,j}$ by replacing it with the closest entry $z_k$ in the learnable codebook $\mathcal{Z}=\left\{z_{k}\right\}_{k=1}^{K} \subset \mathbb{R}^{n_{z}}$, which can be formulated as:
\begin{equation}\label{equ:quantification}
    z_{i,j}=\mathbf{q}(\hat{z}_{i,j}):=\left(\underset{z_{k} \in \mathcal{Z}}{\arg \min }\left\|\hat{z}_{i,j}-z_{k}\right\|\right).
\end{equation}
Finally, the decoder $G$ takes as input the discrete feature map $\mathbf{Z}$ and reconstructs the input image,
\begin{equation}
    \hat{I} = G(\mathbf{Z}) = G(\mathbf{q}(E(\mathbf{I}))).
\end{equation}

Thus, once the VAE is fully trained, the codebook $\mathcal{Z}$ and the generator (i.e., decoder) $G$ can be obtained.

Given the learned VAE and the codebook, each image in the dataset can be converted into a discrete feature map $\mathbf{Z}$ and further be represented as an index sequence $\mathbf{S} \in \{0,...,|\mathcal{Z}|-1\}^{h \times w}$, in which each element represents the entry index in the codebook $\mathcal{Z}$. In the second stage, a sequence generation model (e.g., Transformer) is then used to model the sequence distribution $\mathbf{Pr}(\mathbf{S}|\mathbf{C})$ conditioned on the control signals $\mathbf{C}$.
More specifically, taking as inputs the control signals $\mathbf{C}$ and the preceding predicted tokens $\mathbf{S}_{<i}$, the sequence generation model predicts a conditional distribution $\mathbf{Pr}(s_i|\mathbf{S}_{<i},\mathbf{C})$ for the next token $s_i$. Finally, the predicted index sequence $\mathbf{S}$ is converted to a discrete feature map and fed into the generator $G$ to generate a new image.

\subsection{MaskCLIP: Mask-level Contrastive Language-Image Pretraining}
\label{sec:maskClip}
The learnable codebook in the existing two-stage-based methods~\cite{dalle,zhang2021ufc} struggles to represent visual parts that do not (or only rarely) exist in the training set, largely impairing its expressivity during inference. To address this issue, our ARMANI framework combines each visual code in the codebook with its most relevant word embedding, resulting in a text-aware cross-modal codebook with more distinct semantics.

However, directly combining a visual code with its corresponding word embedding is not trivial, since the visual code and the word embedding are in different feature spaces and the correspondence between these two features is not accessible. To align the visual code and the word embedding and at the same time capture the detailed semantic information of the garment, we propose to leverage the garment segmentation mask, which has been proven useful in vision-language understanding to gain more fine-grained image information~\cite{zhang2021vinvl,lu2019vilbert}.
In particular, ARMANI introduces a Mask-level Contrastive Language-Image Pretraining (MaskCLIP) module to explicitly learn the correspondence between the word embedding and the visual part through the segmentation mask (see Fig.~\ref{3}).
Concretely, during training, given a training batch $\{\mathbf{I}^k,\mathbf{T}^k\}_{k=1}^{b}$, a garment segmentation mask $\mathbf{M}^k$ is first obtained from the garment image $\mathbf{I}^k$ using a mask prediction network to represent the detailed parts of the garment such as button, sleeve, belt, etc. The mask is then encoded as the visual token $\mathbf{\hat{Z}}^k \in \mathbb{R}^{n_1 \times d}$. Similarly, the text sample $\mathbf{T}^k$ is encoded producing the text token $\mathbf{\hat{X}}^k \in \mathbb{R}^{n_2 \times d}$.
We then compute the similarity between each entry $z_i^k$ in $\mathbf{\hat{Z}}^k$ and all the text tokens $\mathbf{\hat{X}}^k$ and let the largest similarity represent the token-wise similarity between the visual token $z_i^k$ and the whole text:
\begin{equation}
\max _{0 \leq j<n_{2}}({z_i^k}^{\top}\cdot x_j^k).
\end{equation}
The similarity between a mask and a text is then computed as the average over the token-wise similarity for the given mask. Thus, the similarity of $\mathbf{M}^p$ to $\mathbf{T}^q$ can be formulated as:
\begin{equation}
    sim^{p, q}_{M}\left(\mathbf{M}^p,\mathbf{T}^q\right)=\frac{1}{n_{1}} \sum_{i=1}^{n_{1}}({z_i^p}^{\top} \cdot x_{j^*}^q),
\end{equation}
where $j^*=\arg \max _{0 \leq j<n_{2}}({z_i^p}^{\top} \cdot x_j^q)$. Similarly, the similarity of $\mathbf{T}^q$ to $\mathbf{M}^p$ can be formulated as:
\begin{equation}
    sim^{p, q}_{T}\left(\mathbf{M}^p,\mathbf{T}^q\right)=\frac{1}{n_{2}} \sum_{j=1}^{n_{2}}({z_{i*}^p}^{\top} \cdot x_{j}^q),
\end{equation}
where $i^*=\arg \max _{0 \leq i<n_{1}}({z_i^p}^{\top} \cdot x_j^q)$. Based on the definition of the token-wise similarity between the image and the text, the image-to-text token-wise contrastive loss for a training batch and its text-to-image counterpart can be calculated as:
\begin{equation}
    \mathcal{L}^{p}_{M}\left(\mathbf{M}^p,\left\{\mathbf{T}^q\right\}_{q=1}^{b}\right)=-\frac{1}{b} \log \frac{\exp \left(sim^{p,p}_{M}\right)}{\sum_{q} \exp \left(sim^{p, q}_{M}\right)},
\end{equation}

\begin{equation}
    \mathcal{L}^{q}_{T}\left(\mathbf{T}^q,\left\{\mathbf{M}^p\right\}_{p=1}^{b}\right)=-\frac{1}{b} \log \frac{\exp \left(sim^{q,q}_{T}\right)}{\sum_{p} \exp \left(sim^{p, q}_{T}\right)}.
\end{equation}

\noindent The total loss of our MaskCLIP module can be formulated as:
\begin{equation}
    \mathcal{L}_{MaskCLIP} = \frac{1}{2} \sum_{k=1}^{b}\left(\mathcal{L}^{k}_{M}+\mathcal{L}^{k}_{T}\right).
\end{equation}

\subsection{Cross-modal Codebook}
\label{sec:codebook}
After training MaskCLIP, we learn the cross-modal codebook by injecting the relevant word embedding into each visual code. More specifically, when learning the codebook, we first get the mask label of the input image $\mathbf{M}$ and its related text $\mathbf{T}$. For each visual part token $\hat{z}_i$ in $\mathbf{\hat{Z}}$, we then obtain its related text token $\hat{x}_j$ according to the similarity of the mask token and the text tokens, namely,
\begin{equation}
    \hat{x}_j = \arg \max _{\hat{x}_k \in \mathbf{\hat{X}} }({\hat{z}_i}^{\top} \cdot \hat{x}_k).
\end{equation}

A cross-modal local feature $\Tilde{z}_i$, which incorporates the visual information as well as the fine-grain text information, can then be obtained by averaging the visual token $\hat{z}_i$ and its corresponding text token $\hat{x}_j$. Finally, all cross-modal features $\Tilde{z}_i$ are quantized using Eq.~\ref{equ:quantification}, and form the discrete feature map $\mathbf{Z}$, which is further sent to the generator to reconstruct the input image $\mathbf{I}$.
In addition, we use the proposed MaskCLIP loss to ensure that the cross-modal codebook captures semantically accurate information by optimizing a symmetric cross entropy loss over the similarities of an image and text embedding batch.
This results in the semantic loss
\begin{equation}
    \mathcal{L}_{semantic} =1-D(M(G(\hat{z}_i)),T).
\end{equation}
where $D$ is the similarity between the MaskCLIP embeddings of its two arguments, $T$ is the text description of the picture, and $M$ is the mask prediction model.

The total loss in the first stage can thus be formulated as:
\begin{equation}
    \mathcal{L}_{stage_1} = \lambda_1\mathcal{L}_{VQ-GAN}+\lambda_2\mathcal{L}_{semanitic},
\end{equation}
where $\mathcal{L}_{VQ-GAN}$ denotes the training losses used in VQ-GAN~\cite{esser2021taming}.
\begin{figure*}[htbp] 
\centering 
\includegraphics[width=0.95\textwidth]{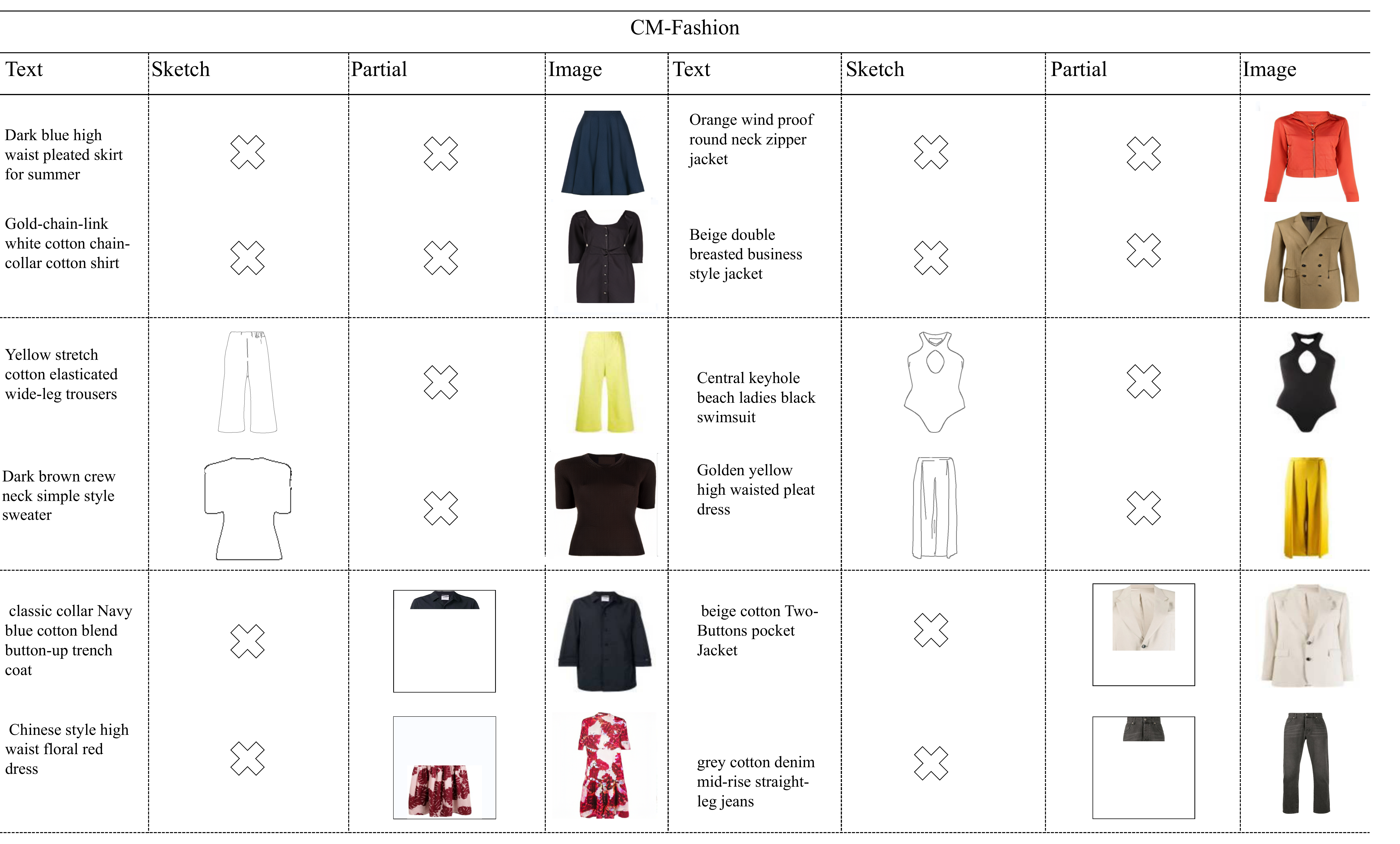}
\vspace{-5mm}
\caption{Qualitative image generation results of ARMANI from text or a combination of text, sketch, and partial images. } 

\vspace{-4mm}
\label{4} 
\end{figure*}

\subsection{Cross-modal Transformer}%model
\label{sec:transformer}

In contrast to existing cross-modal image synthesis approaches~\cite{esser2021taming,styleclip,tan2021cross,kenan2020learning,zhang2021cross} which can only accomplish text-to-image synthesis or image-to-image synthesis, our ARMANI framework is a versatile model that can handle diverse control signals within a unified network.
The main challenge of such a unified model is how to represent diverse control signals in a uniform way. 
To address this challenge, ARMANI introduces a cross-modal Transformer to jointly take diverse control signals as input and predict a sequence of discrete tokens that are used to generated the image (see Fig.~3).
As opposed to CNN-based approaches, Transformers are capable of handling this task, since data from various modalities (e.g., images, text, audio, etc.) can be transferred into the same representation, the token sequence. Besides, since Transformers predict the next token in an autoregressive manner, according to the previous token sequence or the condition signals, we can easily add a new control signal into the model by prepending the token sequence of the new signal to the input tokens.

Specifically, the control signals in our ARMANI framework are comprised of pure text, sketches, and partial images, in which the pure text is directly obtained from a cross-modal dataset, while sketches and partial images can be obtained by applying the Canny edge detection algorithm~\cite{canny1986computational} and extracting a random crop of the RGB image, respectively. 

The control signals are first converted into text and image token sequences, where the text token sequence is obtained directly by a Transformer-based encoder. For the image token sequence, we instead learn a codebook for each type of signal and quantize each signal into a discrete image token sequence. The different token sequences are combined and a special token [SEP] is used to indicate the separation between the modalities. The cross-modal Transformer then predicts the token sequence and feeds it to the image decoder to synthesize the image. 

During training, the cross-modal Transformer is trained by maximizing the log-likelihood of the generated token sequence conditioned on various control signals by minimizing the following training loss:
\begin{equation}
    \mathcal{L}_{\text {Transformer }}=\mathbb{E}_{x \sim p(x)}[-\log p(s|c)],
\end{equation}
\begin{equation}
    p(s \mid c)=\prod_{i} p\left(s_{i} \mid s_{<i}, c\right),
\end{equation}
where $s$ represents the predicted token sequence while $c$ denotes the conditional token sequence. To further ensure that the generated image represents the text, we apply our semantic loss $\mathcal{L}_{semanitic}$ for the token prediction to improve the ability to capture the relation between text and image. Finally, the total loss in the second stage can thus be formulated as:
\begin{equation}
    \mathcal{L}_{stage2} = \mathcal{L}_{Transformer}+\lambda_3\mathcal{L}_{semanitic}.
    \vspace{-3mm}
\end{equation}

\begin{figure}[tbp]
\centering 
\includegraphics[width=0.95\linewidth]{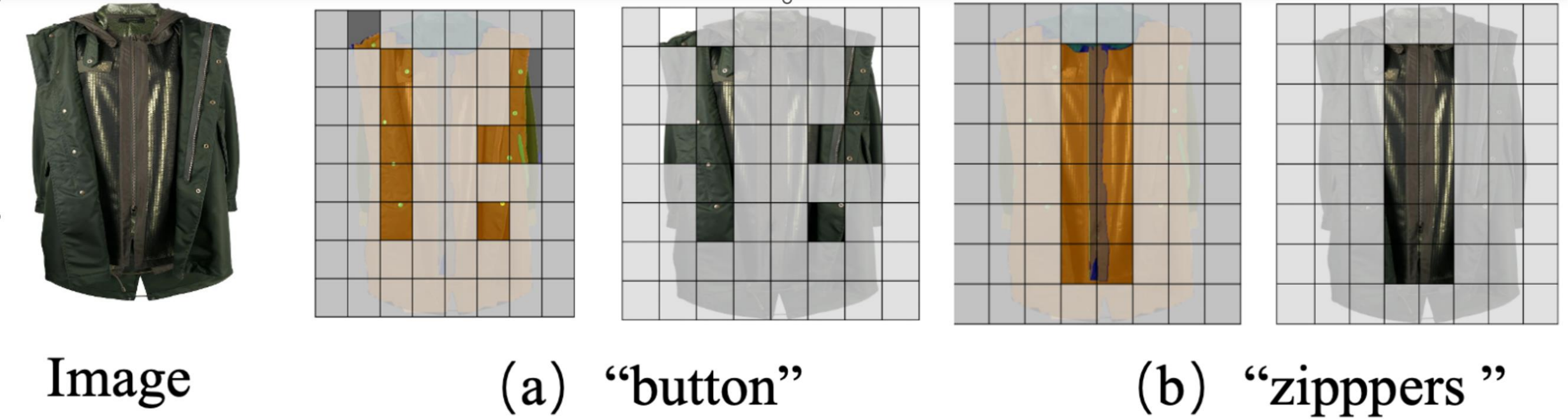} 
\vspace{-1mm}
\caption{Visualizations of the MaskCLIP alignment.} 
\vspace{-5mm}
\label{5}
\end{figure}

\section{Experiments}

\noindent\textbf{Datasets.}
Fashion design research has been limited by the lack of a high-quality garment dataset, as previous approaches like M6-UFC~\cite{zhang2021ufc} have not released their dataset for commercial reasons. Further, existing openly available fashion datasets such as DeepFashion~\cite{DeepFashion} lack detailed text descriptions of the garments.
We, therefore, collect a new dataset from Farfetch for the task of cross-modal fashion synthesis, which we call CM-Fashion. Specifically, our CM-Fashion dataset contains over 500,000 images from a variety of garment categories (e.g., t-shirts, jackets, dresses, pants, etc.) and includes detailed descriptions. 
We will release the dataset at \url{https://github.com/Harvey594/ARMANI} to facilitate the development of fashion design approaches.
%and have included example samples in the appendix. 
CM-Fashion comprises the first open source data set with text descriptions in the field of fashion design and we believe that it will pave the way towards further development of methods tailored towards cross-modal fashion design such as ARMANI. Details of the dataset are provided in the appendix.

\noindent\textbf{Baselines.}
We verify the effectiveness of ARMANI on three tasks, namely, the Text-to-Image, Text+Sketch-to-Image, and Text+Partial-to-Image tasks. To validate the benefits of ARMANI over existing methods, we compare to three cross-modal synthesis works, namely, TediGAN~\cite{xia2020}, DALL·E~\cite{dalle}, and Cogview~\cite{ding2021}. Since DALL·E~\cite{dalle} also leverages
the two-stage paradigm, we ensure a fair comparison by choosing VQGAN~\cite{esser2021taming} instead of VQVAE2~\cite{razavi2019generating} in the first stage and extend their framework to the Text+Sketch-to-Image and Text+Partial-to-Image tasks with a cross-modal Transformer. For TediGAN~\cite{xia2020}, which is a GAN-based approach, we instead use their proposed complex instance-level optimization scheme and use the official implementation of Cogview~\cite{ding2021}. %\footnote{Note, we are unable to compare ARMANI with the relevant work M6-UFC~\cite{zhang2021ufc}, since neither the code nor the pre-trained models or their fashion dataset, M2C-Fashion, have been released.}

\noindent\textbf{Implementation Details.}
Our proposed ARMANI model is implemented using PyTorch and is trained on 8 Tesla V100 GPUs. For the mask prediction network, we use Pointrend of Detecton2~\cite{wu2019detectron2} as our segmentation network, which is trained on a private dataset consisting of 45,000 labeled images. The final predicted segmentation has 15 labels, each label representing a garment part such as pocket, waistband, buttons, etc. The full label list and corresponding visualizations are provided in the appendix. To facilitate reproducibility, the trained mask prediction model will be released together with the code of ARMANI and the CM-Fashion dataset upon acceptance. 
During MaskCLIP training, the text and image encoders are the same as the ones used in CLIP~\cite{radford2021learning} and consist of a Transformer with 12-layers, 8 attention heads, 512 hidden state size, and a total of 63M parameters. During the first training stage, we convert images of size 256*256 into a sequence of 16*16 codes and set the codebook size to 1024. In the second stage, we use a GPT2-medium~\cite{radford2019language} architecture with 16 attention heads, embedding dimensionality of 1024 and 16 transformer blocks, resulting in 407M parameters. As for the hyper-parameters of the first stage loss $\lambda_1$, $\lambda_2$, and $\lambda_3$ are set to 0.9, 0.1, and 2, respectively. Additional implementation details are provided in the appendix.%\commentZ{delete footnoe}%\footnote{Additional implementation details of the mask prediction network and our model are provided in the supplementary material.} 

\begin{figure*}[htbp] 
\centering 
\includegraphics[width=1.03\linewidth]{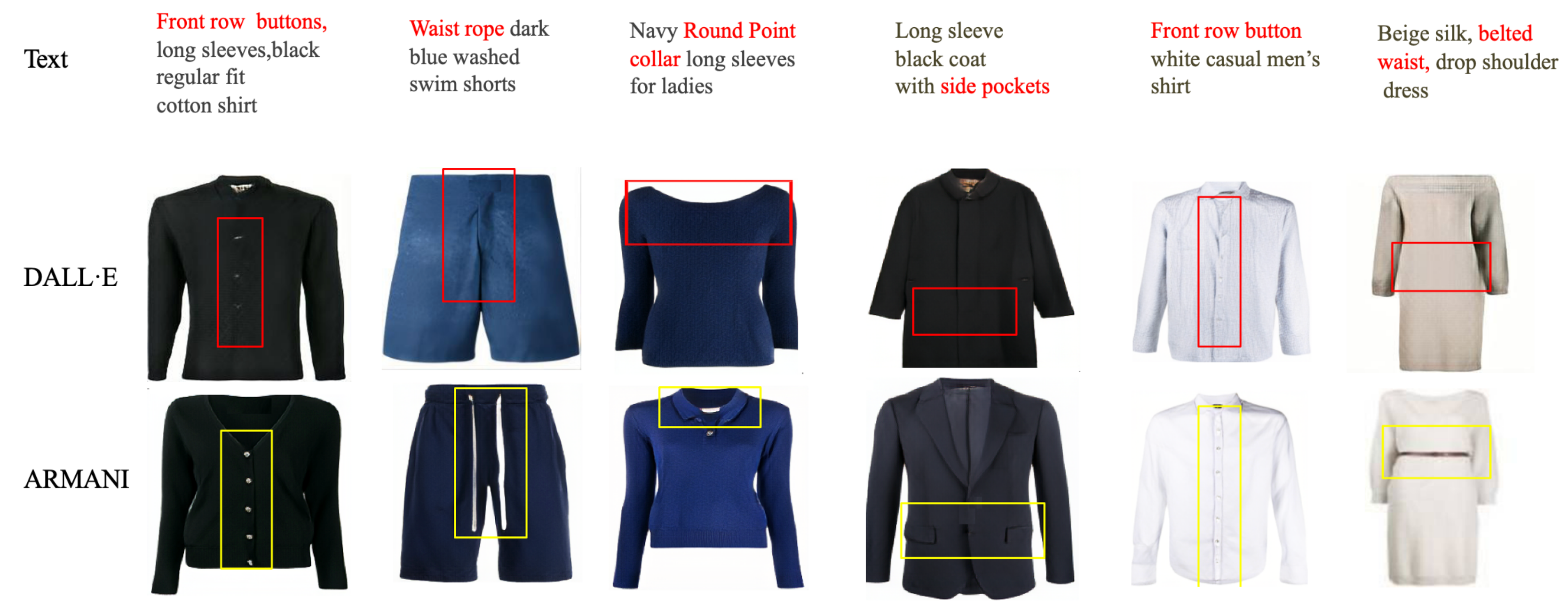} 
\caption{A comparison between ARMANI and DALL·E~\cite{dalle} on the Text-to-Image synthesis task for some difficult examples that require the precise generation of fine-grained details. The boxes are used to highlight specific areas that should contain the elements highlighted in the text.}

\vspace{-4mm}
\label{6} 
\end{figure*}

\begin{table*}[t]
\centering

 \begin{tabular}{l c c c c c c c c c c c c c c c }
   \toprule
   \multicolumn{2}{c}{Task}& & \multicolumn{3}{c}{Text-to-Image}& &\multicolumn{3}{c}{Text+Sketch-to-image} &&\multicolumn{3}{c}{Text+Partial-to-Image} \\
   \cmidrule{1-2} \cmidrule{4-6} \cmidrule{8-10} \cmidrule{12-14}
   \multicolumn{2}{c}{Metric} & & FID $\downarrow$ & IS $\uparrow$ & CLIPScore $\uparrow$ &&FID $\downarrow$ & IS $\uparrow$ & CLIPScore $\uparrow$ && FID $\downarrow$ & IS $\uparrow$ & CLIPScore$\uparrow$ \\
   \cmidrule{1-2}  \cmidrule{4-6} \cmidrule{8-10} \cmidrule{12-14}
   \multicolumn{2}{c}{TediGAN~\cite{xia2020}} & & 27.378 & 18.46  & 0.5587 &  &——& —— & —— && —— & ——  & —— \\
   \multicolumn{2}{c}{Cogview~\cite{ding2021}} & & \textbf{12.198} & 23.99  & 0.6572 &  &——& —— & —— && —— & ——  & —— \\
   \multicolumn{2}{c}{DALL·E~\cite{dalle}} & & 13.249 & 20.33  & 0.6423 & &12.578 & 24.98 & 0.7028&& 13.287 & 22.99  & 0.6812 \\
   \cmidrule{1-2} \cmidrule{4-6} \cmidrule{8-10} \cmidrule{12-14}
   \multicolumn{2}{c}{ARMANI (Ours)} & & 12.336 & \textbf{24.32}  & \textbf{0.6988} & & \textbf{11.882} & \textbf{25.29} & \textbf{0.7433} & & \textbf{11.741} & \textbf{24.03}  & \textbf{0.7189} \\
   \bottomrule
 \end{tabular}
 \caption{Comparisons with TediGAN~\cite{xia2020}, Cogview~\cite{ding2021}, and DALL·E~\cite{dalle} for the three tasks of the CM-Fashion dataset. $\downarrow$ means the lower the better, while $\uparrow$ means the opposite. Note, TediGAN~\cite{xia2020} and Cogview~\cite{ding2021} do not support the latter two tasks.}
 \label{1.1}
 \vspace{-8mm}
 \end{table*}

\subsection{Visualization of MaskCLIP}
In this section, we visualize MaskCLIP’s capability of capturing fine-grained semantically-accurate cross-modal correspondence using the proposed word-part alignment.

\noindent\textbf{Visualization Strategy.}
The word-part alignment is performed based on the token-wise similarity between the mask patches and the textual tokens. Specifically, for the text, the visual token with the largest similarity to it is considered as its predicted label. Note that one garment part may be tokenized to more than one mask token. For an image and its mask, MaskCLIP predicts the patch that is most relevant to the given text. We highlight the mask patch and the image patch corresponding to the text, while the others are marked in white.

\noindent\textbf{Visualization result.}
Fig.~\ref{5} shows the word-part alignment results for MaskCLIP. The first image shows the correspondence between the text and the mask, and the second image shows the correspondence between the original image and the mask. As can be seen, MaskCLIP accurately predicts the mask token that is most relevant to a given text.

\subsection{Comparison With State-Of-The-Art Methods}

\noindent\textbf{Qualitative Results.}
We first present qualitative results for ARMANI in Fig.~\ref{4} for all three tasks, before comparing to DALL·E~\cite{dalle}.
For the Text-to-Image synthesis task (Fig.~\ref{6}), ARMANI can synthesize realistic fashion images that comply with the textual description, while DALL·E~\cite{dalle} generates garment images that comply with the overall content of the textual description, but tends to neglect fine-grained information in the input text. Moreover, DALL·E fails to synthesize precise details such as buttons and small logos (red box in Fig.~\ref{6}).
ARMANI on the other hand is capable of recovering the fine-grain details corresponding to each word in the input text due to the explicit introduction of the text tokens into the codebook and accurate sequence prediction.
We observe similar behavior for the other two tasks and provide the qualitative results in the appendix.

\noindent\textbf{Quantitative Results.}
We apply FID~\cite{heusel2017gans} and IS~\cite{salimans2016improved} to measure the quality of the synthesized images. Further, we use CLIPScore~\cite{hessel2021CLIPScore} to measure the relevance of the text to a given image. A higher CLIPScore~\cite{hessel2021CLIPScore} indicates that the text is more relevant to the image. In addition, we designed a human evaluation to compare the baselines and our generation results. A higher human evaluation score indicates that a larger fraction of participants preferred the results of a given method.\footnote{Details on the human evaluation are provided in the appendix.}
As reported in Table \ref{1.1}, our ARMANI model outperforms the baselines TediGAN~\cite{xia2020}, Cogview~\cite{ding2021}, and DALL·E~\cite{dalle} in most cases by a large margin, mostly obtaining the lowest FID~\cite{heusel2017gans} and IS~\cite{salimans2016improved} scores and the highest CLIPScore~\cite{hessel2021CLIPScore} on the three tasks. Note, here DALL·E~\cite{dalle} has been extended to the Text+Partial-to-Image and Text+Sketch-to-Image synthesis tasks.

\begin{figure}[tbp]
\centering 
\includegraphics[width=\linewidth]{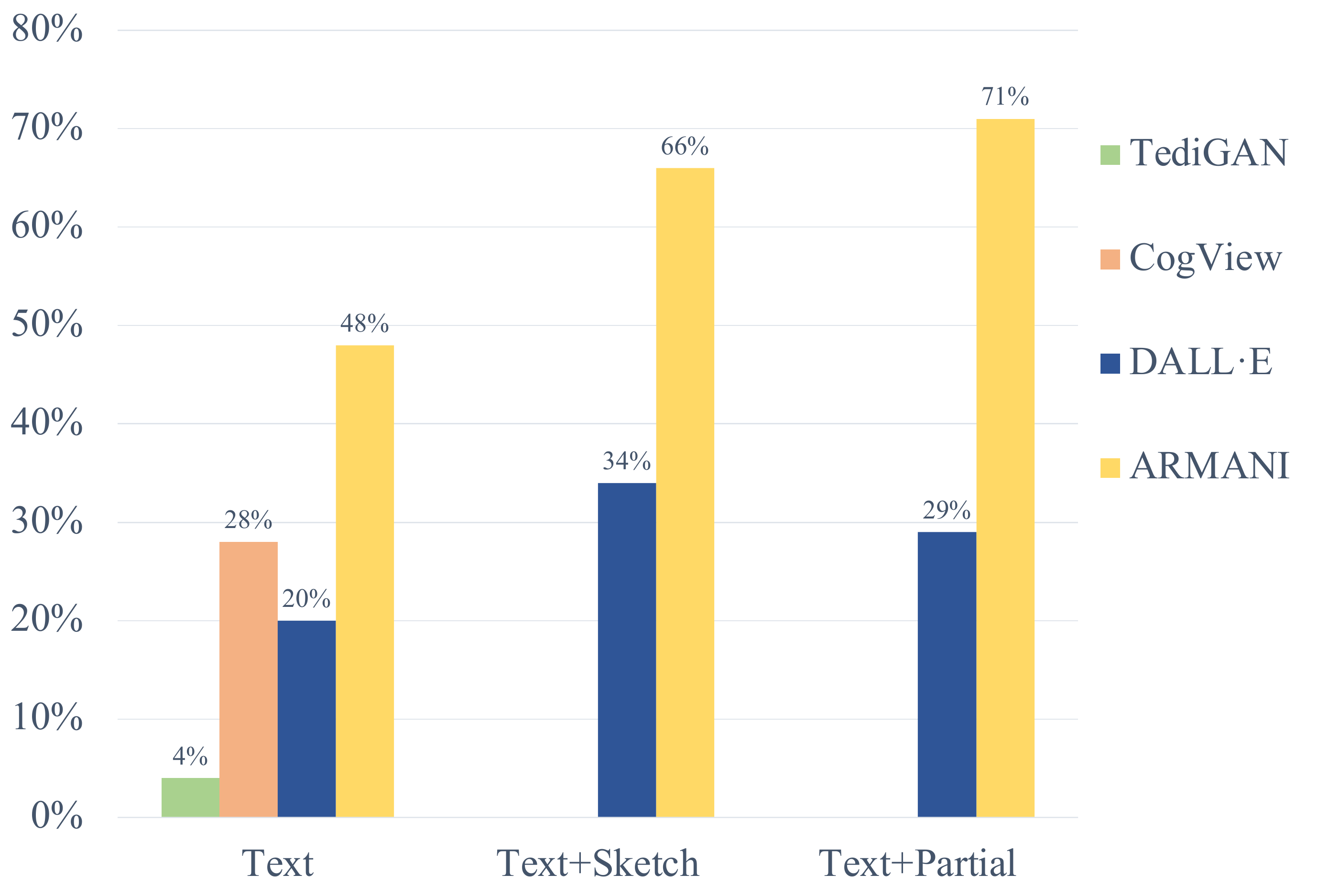} 
\caption{Comparison of the human evaluation results of TediGAN, Cogview, and DALL·E for the three tasks of interest.}
\label{7}
\vspace{-4mm}
\end{figure}

As the corresponding ground truth is not available when generating images based on text conditions, we perform a human evaluation study in order to assess the image quality and its relevance to the text jointly.
We observe from Fig.~\ref{7} that ARMANI's 
results are preferred according to the human evaluation on all the three tasks and we also observe that, compared to the machine evaluation, the human evaluation indicates a larger difference among the models.

\begin{figure}[tbp]
\centering 
\includegraphics[width=\linewidth]{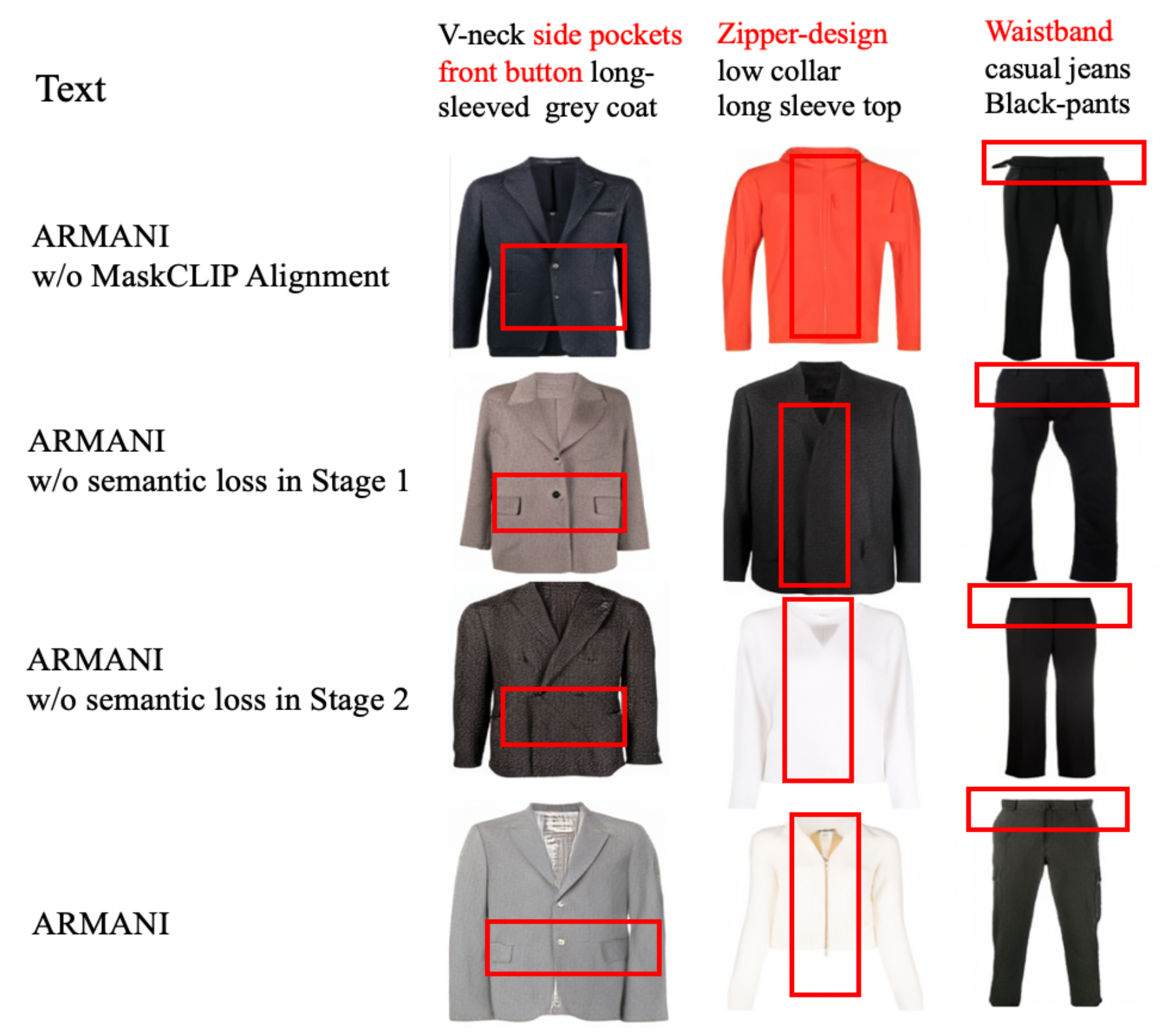} 
\caption{Qualitative comparison of our methods w/o MaskCLIP alignment, w/o semantic loss in the first stage, w/o semantic loss in the second stage, and our full version.}
\vspace{-6mm}
\label{8}
\end{figure}

\subsection{Ablation Study}

To validate the effectiveness of the semantically accurate alignment of our proposed MaskCLIP, we perform an ablation study, where we compare ARMANI with a version of ARMANI without the MaskCLIP module. We compare these models on all three tasks and observe that the full ARMANI model obtains higher CLIPscore~\cite{hessel2021CLIPScore}. 
This illustrates that the generated results are more relevant to the text description, thus validating the effect of MaskCLIP in enabling a semantically accurate cross-modal codebook. To validate the effectiveness of the semantic loss, we perform another ablation study, where we compare ARMANI to two versions of ARMANI, one without the semantic loss in the first stage and one without the semantic loss in the second stage. Again, we compare the CLIPScore~\cite{hessel2021CLIPScore} for all three tasks and observe that the CLIPScore worsen as we remove the semantic losses, indicating that the semantic loss plays a significant role for the generation.
QWe further visualize qualitative results of these ablation studies on the Text-to-Image task in Fig.~\ref{8}. These results further demonstrate the validity of our method. In particular, we observe that if MaskCLIP alignment is removed, the codebook's ability to express garment parts will be weakened. While the model can still leverage the garment part information via the semantic loss in the second stage, the generation results are worse. If we instead remove the semantic loss in the first stage, we observe that the information obtained from the MaskCLIP alignment will not be retained in the codebook, again worsening the generation ability for the garment details. Finally, if the semantic loss is removed in the second stage, it will lead to the garment details being ignored in the code sequence prediction of the generated picture and some parts will be ignored and not generated. Additional qualitative results for the other two tasks are provided in the appendix.
We further show quantitative results on all three tasks in Table~\ref{tab:ablation} to evaluate the effectiveness of our model components and observe that the absence of each component leads to a decrease in CLIPScore.

%Finally, we verify that also with single signal inputs such as in the Sketch-to-Image or Partial-to-Image tasks, our method can generate details due to our cross-modal codebook (see Fig.~\ref{10}).

\begin{table}[t]
\centering
\begin{tabular}{c c c c c c c c c c c c c}
  \toprule
  \multicolumn{3}{c}{Model Setting}& \multicolumn{3}{c}{CLIPScore$\uparrow$}&\\
  \toprule
   MaskCLIP & $L_{stage1}$ & $L_{stage2}$ && Task 1 & Task 2 & Task 3&\\
   \midrule
   \XSolid&\Checkmark&\Checkmark && 0.6779&0.7246&0.7003&\\
   \Checkmark&\XSolid & \Checkmark && 0.6764& 0.7151& 0.7011&\\
   \Checkmark&\Checkmark&\XSolid && 0.6617 &0.7031& 0.6931 &\\
   \midrule
   \Checkmark&\Checkmark&\Checkmark && \textbf{0.6988}&\textbf{0.7433}&\textbf{0.7189}&\\
   \bottomrule
  
\end{tabular}
\caption{Quantitative results of our ablation studies. Task 1, Task 2, and Task 3, refer to the Text-to-Image, Text+Sketch-to-Image, and Text+Partial-to-Image tasks, respectively.}
\label{tab:ablation}
\vspace{-10mm}
\end{table}

\section{Conclusion}
We proposed a pARt-level garMent-text Alignment for uNIfied cross-modal fashion design (ARMANI) that can leverage multi-modal control signals to accomplish diverse fashion manipulations. Empowered by its elaborately designed MaskCLIP module, which combines visual and textual features in a semantically accurate manner, ARMANI is able to learn a powerful cross-modal codebook and generate fine-grained photo-realistic images guided by text, sketches, or partial images. Experimental results highlight ARMANI's ability to capture fine-grained semantic information, outperforming previous methods on the cross-model image synthesis task. Finally, we believe that both ARMANI and our newly released dataset CM-Fashion will inspire research on developing versatile cross-modal fashion image synthesis approaches.

\begin{acks}
This work was supported in part by National Key R\&D Program of China under Grant No. 2020AAA0109700, National Natural Science Foundation of China (NSFC) under Grant No. U19A2073, No. 62103454, Guangdong Province Basic and Applied Basic Research (Regional Joint Fund-Key) Grant No.2019B1515120039,
No.2019A151
5110680, Guangdong Outstanding Youth Fund (Grant No.2021B1515
020061), the Shenzhen Municipal Basic Research Project for Natural Science Foundation  (Grant No.JCYJ20190806143408992), Shenzhen Fundamental Research Program (Project No. JCYJ20190807154211365 ) and CAAI-Huawei MindSpore Open Fund. 
We thank MindSpore (https://www.mindspore.cn/), which is a new deep learning computing framework, for the partial support of this work.
%We thank MindSpore (https://www.mindspore.cn/) for the partial support of this work, which is a new deep learning computing framwork.
\end{acks}

\bibliographystyle{ACM-Reference-Format}
\bibliography{sample-base}
\clearpage

\appendix

\section{Dataset Details}
\label{app:details}

The newly collected CM-Fashion dataset consists of $521,068$ garment images and has the following two key properties: 1) CM-Fashion covers most of the regular garment categories, including pants, dresses, jackets, etc. Further, our dataset contains information about gender and target age. Each category has also been divided into garments for men, women, and kids.
2) The images in the CM-Fashion dataset are accompanied by detailed text descriptions and sketches. The sketches are binary pictures that have been obtained via the Canny edge detection algorithm~\cite{canny1986computational}. Note, the partial images mentioned in our paper are the result of a random crop (keeping $25\%$ of the original area) and therefore not included explicitly in the dataset. %\commentQ{Maybe include details for the random cropping?}\commentZ{it's just a function randomcrop in pytorch,The area is 1/4 of original image,I don't know how to describe this}
After collecting the raw images, data pre-processing is required to exclude invalid data. We removed images whose text description was either too long (longer than 400 words) or missing, and images that contained more than one garment.
Finally, our dataset is divided into a train set containing 468960 images (90\%) and a test set with 52108 images (10\%).
The images in CM-Fashion are crawled from the E-commerce website Farfetch.\footnote{https://www.farfetch.cn/uk/} 
\begin{figure}[tbp]
\centering
\includegraphics[width=0.95\linewidth]{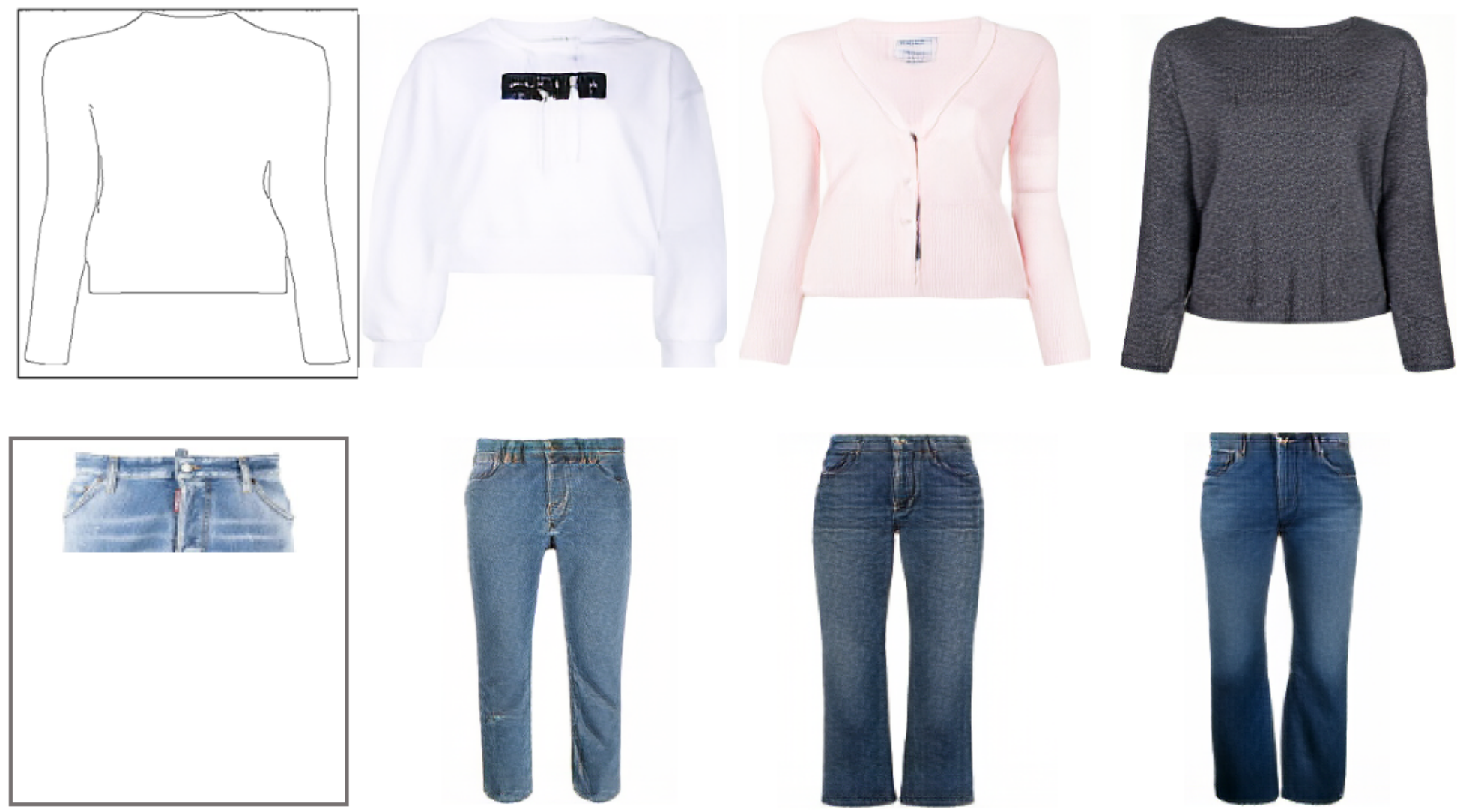} 
\caption{Qualitative image generation results of ARMANI from a sketch or a partial image without text.}
\label{2222}
\end{figure}

\begin{figure*}[htbp] 
\centering 
\includegraphics[width=0.95\textwidth]{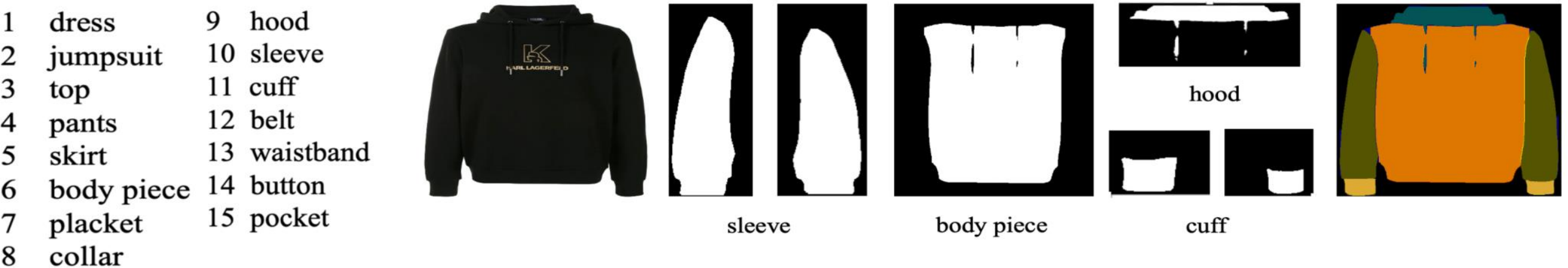} 
\caption{A list over the 15 label categories in the garment segmentation and a garment example with segmentation masks.} 
\label{15} 
\end{figure*}

\section{Additional Implementation Details}
While most of the implementation details have been described in the main paper, we provide the remaining details here. For the mask prediction network, we use PointRend of Detecton2~\cite{wu2019detectron2} as our segmentation network, to obtain the garment segmentation result. There are 15 different labels for the garment segmentation (see Fig~\ref{15}). In the training step of MaskClip, the text encoder consists of a Transformer with 12-layers, 8 attention heads, and a hidden state of dimensionality 512. For the image encoder, we use a ViT-B/16~\cite{kim2021vilt} and for the implementation of the contrastive loss, following CLIP~\cite{radford2021learning}. 
In the first stage, we trained the encoder, the cross-modal codebook, and the decoder for 20 epochs, which took approximately five days on 4 Tesla V100 GPUs with a batchsize of 8. During the second stage,
we use the GPT2 [4] architecture. We trained the model for 10 epochs which took almost four days on 4 Tesla V100 GPUs with a batchsize of 8. The learning rate for the first stage was set to $6 \times 10^{-6}$, while it was set to $8 \times 10^{-4}$ for the second stage.

\section{Human Evaluation Details}
For the human evaluation, we separately design three questionnaires for our ARMANI and the baseline methods, one for each of the investigated settings (Text-to-Image, Text+Sketch-to-Image, and Text+Partial-to-
Image tasks). Each questionnaire is composed of 100 tasks, where the results for the different methods are presented in random order and where the volunteers are asked to pick the result that looks most realistic and relevant to the provided information in the task.
Before the start of the human evaluation, we first invite five volunteers to accomplish the questionnaires in a serious manner to test the time required to complete the questionnaires. During the evaluation, for a particular questionnaire, we then invite 100 random volunteers, who are asked to spend at least 6 seconds to accomplish each task in the questionnaire.

\section{Additional Results}
\textbf{Additional ablation study.}
We verify that also with single signal inputs such as in the Sketch-to-Image or Partial-to-Image tasks, ARMANI can generate details due to our cross-modal codebook (see Fig.~\ref{2222})

\begin{figure*}[htbp] 
\centering 
\includegraphics[width=\textwidth]{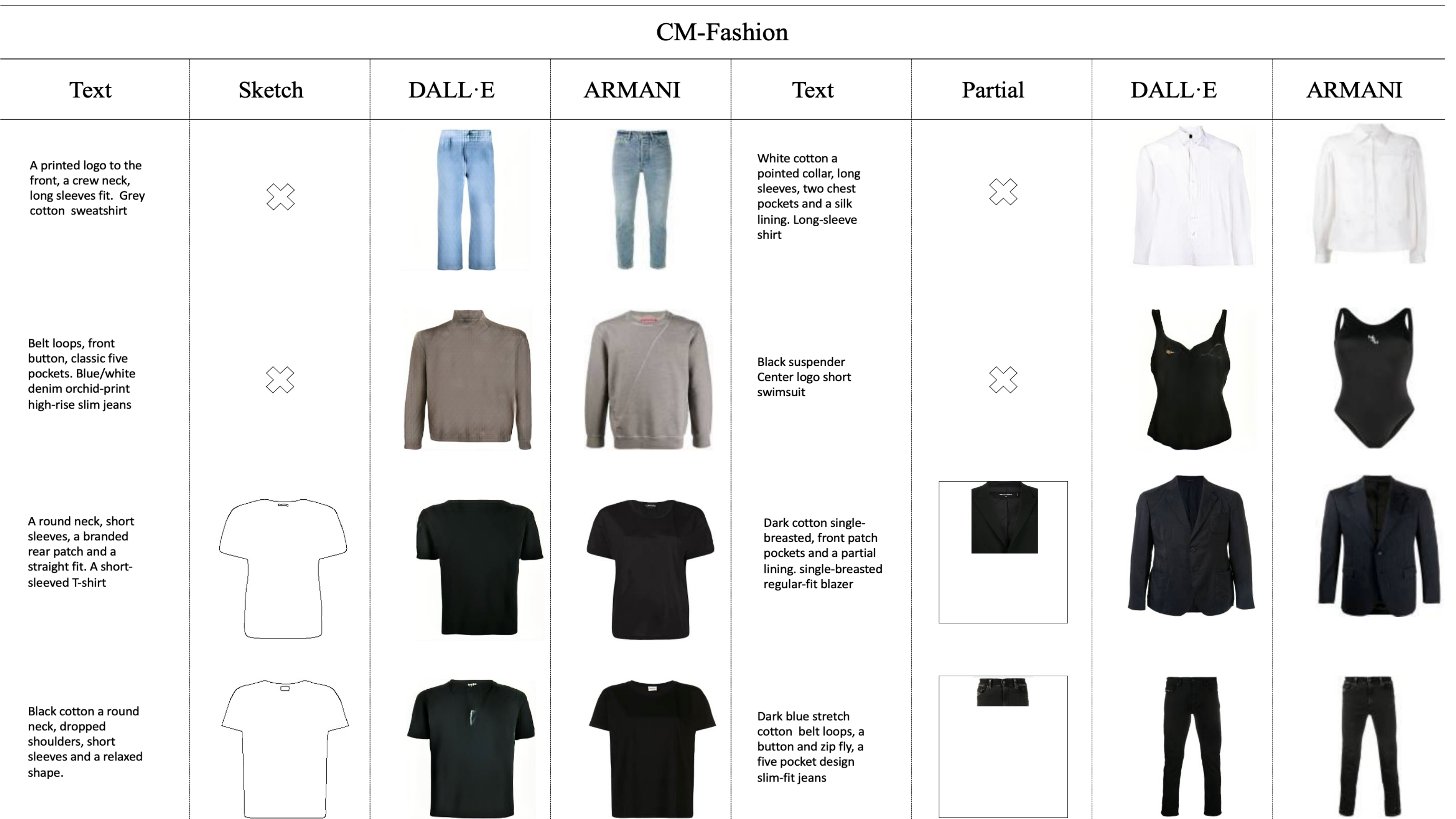}
\caption{Qualitative comparison of ARMANI and  DALL·E~\cite{dalle} on the CM-Fashion dataset. Please zoom in for more details.} 
\label{1111} 
\end{figure*}

\noindent\textbf{Visual Comparison to DALL·E on the CM-Fashion Dataset.}
Fig.~\ref{1111} displays additional qualitative comparisons of ARMANI and DALL·E~\cite{dalle} on the CM-Fashion dataset.

\begin{figure*}[htbp] 
\centering 
\includegraphics[width=\textwidth]{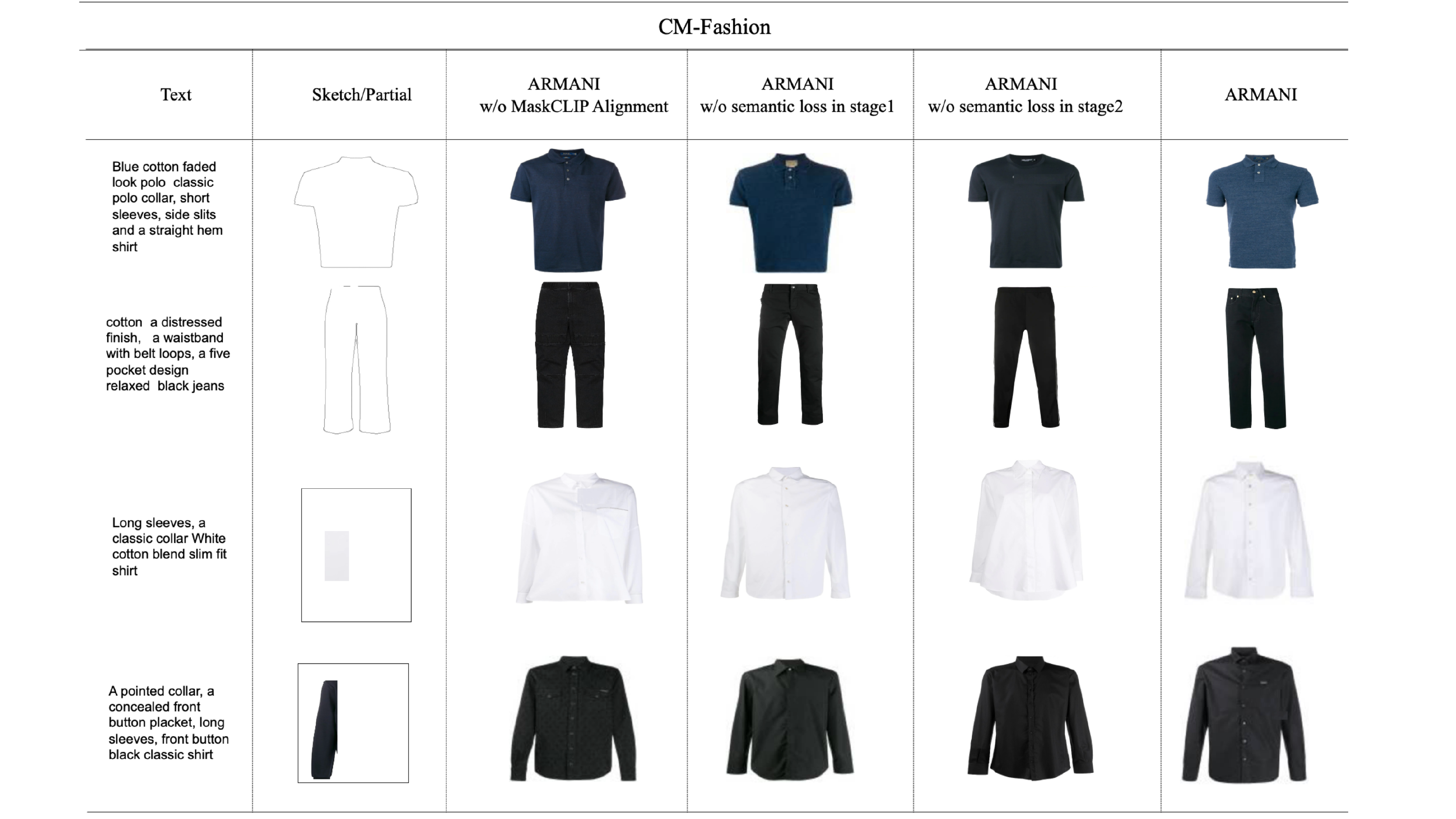}
\caption{Qualitative comparison of our methods w/o MaskCLIP alignment, w/o semantic loss in the first stage, w/o semantic loss in the second stage, and our full version. Please zoom in for more details.} 
\label{111} 
\end{figure*}

\noindent\textbf{Visual Comparison for the ablation study.}
Fig.~\ref{111} displays additional qualitative comparisons of our methods w/o MaskCLIP alignment, w/o semantic loss in the first stage, w/o semantic
loss in the second stage, and our full version.

\section{Limitation}
 Limitations of our approach are: (1) Synthesis results are affected by noisy text. Obtaining completely detailed correspondence to the text becomes challenging when generation is conditioned on complicated and noisy text. (2) Our method is more focused on generating part details. More complex visual information, such as textures, are not directly addressed via our method. However, when the codebook size is increased, we observe that texture generation is also improved.

\end{document}